\documentclass[journal,11pt]{IEEEtran}

\usepackage{cite}

\ifCLASSINFOpdf
   \usepackage[pdftex]{graphicx}
\else
\fi

\usepackage{amsmath}

\usepackage{wrapfig}

\usepackage{graphicx}
\usepackage{subcaption}
\usepackage[latin1]{inputenc}
\usepackage{amssymb}
\usepackage{multirow}
\usepackage{rotating}
\usepackage{picins}
\usepackage{floatflt}

\usepackage{algorithm}
\usepackage{algpseudocode}

\usepackage{booktabs}
\usepackage{color}
\newcommand{\revision}[1]{\textcolor{black}{#1}}

\begin{document}
%
\title{Recalibrating Fully Convolutional Networks with Spatial and Channel `Squeeze \& Excitation' Blocks}

\author{Abhijit Guha Roy, Nassir Navab and Christian Wachinger

\thanks{ 
A. Guha Roy and C. Wachinger are with lab for Artificial Intelligence in Medical Imaging (AI-Med), Child and Adolescent Psychiatry, Ludwig-Maximilians-University M\"{u}nchen, Germany.
A. Guha Roy and N. Navab are with Chair for Computer Aided Medical Procedures (CAMP), Technical University of Munich, Germany. N. Navab is with Chair for Computer Aided Medical Procedures (CAMP), Johns Hopkins University, Baltimore, USA.
}
}

\markboth{Accepted article to appear in IEEE Transactions on Medical Imaging, 2018. Please cite the journal version}
{Guha Roy \MakeLowercase{\textit{et al.}}: Fully Conv Squeeze \& Excitation}

\maketitle

\begin{abstract}
In a wide range of semantic segmentation tasks, fully convolutional neural networks (F-CNNs) have been successfully leveraged to achieve state-of-the-art performance. Architectural innovations of F-CNNs have mainly been on improving spatial encoding or network connectivity to aid gradient flow. In this article, we aim towards an alternate direction of recalibrating the learned feature maps adaptively; boosting meaningful features while suppressing weak ones. 
The recalibration is achieved by simple computational blocks that can be easily integrated in F-CNNs architectures. 
We draw our inspiration from the recently proposed `squeeze \& excitation' (SE) modules for channel recalibration for image classification. Towards this end, we introduce three variants of SE  modules for segmentation, (i) squeezing spatially and exciting channel-wise, (ii) squeezing channel-wise and exciting spatially  and (iii) joint spatial and channel squeeze \& excitation. 
We effectively incorporate the proposed SE blocks in three state-of-the-art F-CNNs and demonstrate a consistent improvement of segmentation accuracy on three challenging benchmark datasets. 
Importantly, SE blocks only lead to a minimal increase in model complexity of about 1.5\%, while the Dice score increases by 4-9\% in the case of U-Net. 
Hence, we believe that SE blocks can be an integral part of future F-CNN architectures.

\end{abstract}


\IEEEpeerreviewmaketitle


\section{Introduction}
\label{sec:intro}
Deep learning based architectures, especially convolutional neural networks (CNN), have become the tool of choice for processing image data, after their immense success in image classification~\cite{alexnet2012,resnet}. 
For image segmentation, fully convolutional neural networks (F-CNNs) have set the benchmark performance in medical imaging~\cite{Unet,Vnet,ecb2017} and computer vision~\cite{longfcn2015,deconvnet2015,segnet,densenet}.
The  basic building block for all these architectures is the convolutional layer, which learns filters capturing local spatial pattern along all the input channels and generates feature maps jointly encoding the spatial and channel information. 
This produces feature maps that form a rich representation of the original input. 
A lot of recent work has aimed at improving the joint encoding of spatial and channel information~\cite{chen2018deeplab,dai2017deformable}, but much less attention has been given towards encoding of spatial and channel-wise patterns \emph{independently}. 
A recent work attempted to address this issue by explicitly modeling the interdependencies between the channels of  feature maps to enhance its representation.
This is accomplished by an architectural component called squeeze \& excitation (SE) block~\cite{SE2017}, which can be seamlessly integrated as an add-on within a CNN.
This SE block factors out the spatial dependency by global average pooling to learn a channel specific descriptor, which is used to rescale the input feature map to highlight only useful channels. As this component `squeezes' along spatial domain and `excites' or reweights along the channels, it is termed as squeeze \& excitation block.
A convolutional network with such SE blocks achieved the best performance in the ILSVRC 2017 image classification competition on the ImageNet dataset, indicating its efficiency~\cite{SE2017}.

In this article, we aim at leveraging the high performance of SE blocks for image classification to image segmentation, by integrating them within F-CNNs. 
We refer to the previously proposed SE block~\cite{SE2017} as channel SE (cSE), because it only excites channel-wise, which has been shown to be highly effective for image classification. The advantage of using such a block is that every intermediate layer has the total receptive field of the input image, due to the global average pooling.
We hypothesize that the pixel-wise spatial information is more informative for fine-grained segmentation tasks of highly complex anatomies, common in medical imaging. 
Hence, we introduce an alternate SE block, which `squeezes' along the channels and `excites' spatially, termed \emph{spatial} SE (sSE). This is complementary to the cSE block, as it does not change the receptive field, but provides spatial attention to focus on certain regions.
Finally, we propose to combine these two blocks in spatial and channel SE blocks (scSE) that recalibrate the feature maps separately  along channel and space, and then combines the output. 
This aggregates the unique properties of each of the blocks and encourages feature maps to be more informative both spatially and channel-wise.
We explore different aggregation strategies for both blocks with respect to the segmentation accuracy.
To the best of our knowledge, this is the first time that \emph{spatial} squeeze \& excitation is proposed for neural networks and the first integration of squeeze \& excitation in F-CNNs.

We integrate the existing (cSE) and proposed (sSE, scSE) SE blocks within three state-of-the-art F-CNN models for image segmentation to demonstrate that SE blocks are a generic network component to boost performance. 
We evaluate the segmentation performance in three challenging medical applications: whole-brain, whole-body  and retinal layer segmentation. 
In whole-brain segmentation,  we automatically parcellate 27 cortical and subcortical structures on magnetic resonance imaging (MRI) T1-weighted brain scans. 
In whole-body segmentation, we label 10 abdominal organs on contrast-enhanced CT scans.
In retinal layer and fluid segmentation, we segment retinal Optical Coherence Tomography (OCT) scans into 7 layers and accumulated fluid in subjects with diabetic macular edema.

This work is an extension of our early work~\cite{roy2018concurrent}, where we further improved  the method, provide details, and added more extensive experiments together with an analysis of SE network dynamics during training.

\revision{
To summarize, the contributions of this article are
\begin{enumerate}
\item  The integration of squeeze \& excitation in F-CNNs for semantic segmentation.
\item The introduction of  channel squeeze and spatial excitation (sSE), to provide attention to challenging spatial regions, aiding fine-grained segmentation.
\item The combination of the two cSE and sSE blocks by an element-wise max-out layer to jointly re-calibrate feature maps both channel-wise and spatially.
\item Experiments on 3 challenging segmentation tasks, integrating the proposed blocks within 4 F-CNN architectures, where a consistent improvement in segmentation is observed within minimal increase in model complexity.
\end{enumerate}
}

\subsection{Related Work}
F-CNN architectures have been extensively used in a wide range of medical image segmentation tasks, providing state-of-the-art performance. One of the seminal F-CNN models, U-Net~\cite{Unet} was proposed for segmenting electron microscope scans. U-Net has an encoder-decoder based structure, separated by a bottleneck layer. Skip connections are included between feature maps of encoder and decoder with similar spatial resolution, to provide more contextual information to the decoder and aiding flow of gradient through the network. It was successfully leveraged for segmentation for multiple modalities of medical imaging. Skip-DeconvNet (SD-Net)~\cite{ecb2017} was introduced, which builds on top of U-Net, modifying the decoding path by using unpooling layers~\cite{deconvnet2015} to promote spatial consistency in the segmentation. It is learned by optimizing a joint loss function of weighted logistic loss and Dice loss, specifically designed to address the challenge of class imbalance, which is very common in medical imaging. SD-Net was successfully used for whole-brain segmentation of MRI scans and retinal layer segmentation task in OCT scans~\cite{roy2017relaynet}.
A more recent architecture introduces dense connectivity within CNNs~\cite{huang2017densely}, to promote feature reusability within layer making representation learning more efficient. This idea was incorporated within F-CNNs by having dense connections within the encoder and decoder blocks, unlike U-Net and SD-Net which uses normal convolutions. Such architectures, termed fully convolutional DenseNet (FC-DenseNet)~\cite{densenet}, further boosted segmentation performance. A variant of this FC-DenseNet, has been used for the task of whole brain segmentation in MRI T1 scans~\cite{roy2018quicknat}. In this article, we select these commonly used F-CNN architectures as references to examine the effectiveness of the SE blocks.


\section{Methods}
\label{sec:method}
Given an input image $\mathbf{I}$, F-CNN approximates a non-linear mapping $\mathbf{F}_{seg}(\cdot)$, which maps $\mathbf{I}$ to a segmentation map $\mathbf{S}$, $\mathbf{F}_{seg}:\mathbf{I}\rightarrow\mathbf{S}$. The function $\mathbf{F}_{seg}(\cdot)$ is a sequence of cascaded functions $\mathbf{F}_{tr}^i(\cdot)$  corresponding to each encoder or decoder block, separated by either a max-pooling (in encoder path) or an upsampling layer (in decoder path). 

Let us represent an intermediate input feature map as $\mathbf{X} \in \mathbb{R}^{H \times W \times C'}$ that passes through an encoder or decoder block $\mathbf{F}_{tr}(\cdot)$ to generate output feature map $\mathbf{U} \in \mathbb{R}^{H \times W \times C}$,  $\mathbf{F}_{tr}:\mathbf{X}\rightarrow\mathbf{U}$.
Here $H$ and $W$ are the spatial height and width, with $C'$ and $C$ being the input and output channels, respectively. The generated $\mathbf{U}$  combines the spatial and channel information of $\mathbf{X}$ through a series of convolutional layers and non-linearities defined by $\mathbf{F}_{tr}(\cdot)$. We place the SE blocks $\mathbf{F}_{SE}(\cdot)$ on $\mathbf{U}$ to recalibrate it to $\hat{\mathbf{U}}$. We propose three different variants of SE blocks, which are detailed next. 

\begin{figure}[h]
\centering
\includegraphics[width=0.45\textwidth]{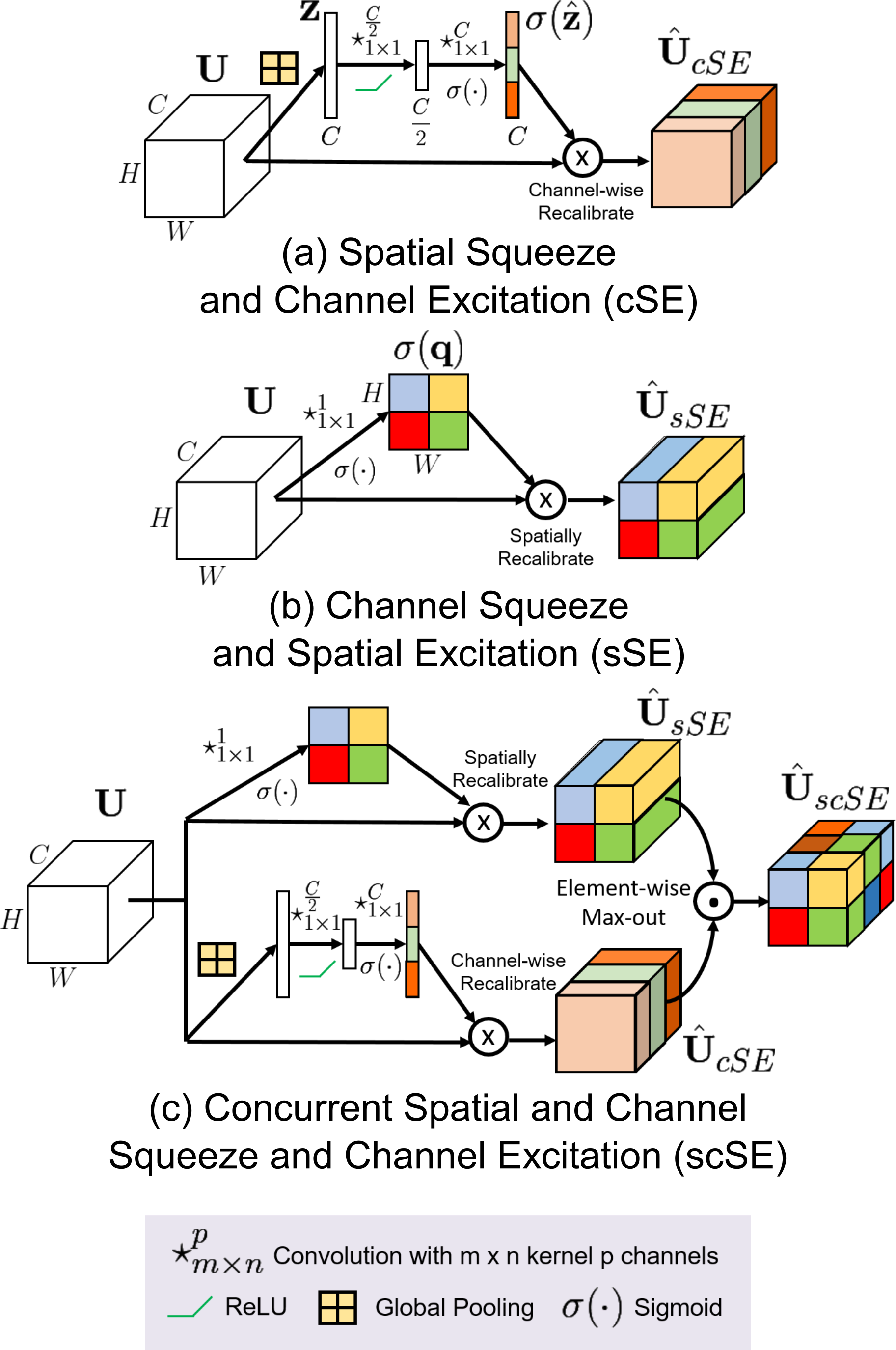}

\caption{Architectural configuration of the cSE, sSE and scSE blocks are shown in (a), (b) and (c), respectively. The input feature map $\mathbf{U}$ is recalibrated for all the cases.
}
\label{fig:GA}
\end{figure}

\noindent
\subsection{Spatial Squeeze and Channel Excitation Block (cSE)}
We describe the spatial squeeze and channel excitation block, which was proposed in~\cite{SE2017}. 
We consider the input feature map $\mathbf{U} = [\mathbf{u}_1, \mathbf{u}_2, \cdots, \mathbf{u}_{C}]$ as a combination of channels $\mathbf{u}_i \in \mathbb{R}^{H \times W}$.
Spatial squeeze is performed by a global average pooling layer, producing vector $\mathbf{z} \in \mathbb{R}^{1 \times 1 \times C}$ with its $k^{th}$ element
\begin{equation}
z_k = \frac{1}{H \times W} \sum_i^H \sum_j^W \mathbf{u}_k (i,j).
\end{equation}

\noindent
This operation embeds the global spatial information in vector $\mathbf{z}$. This vector is transformed to $\hat{\mathbf{z}}=\mathbf{W}_1 (\delta(\mathbf{W}_2 \mathbf{z}))$, \revision{with $\mathbf{W}_1 \in \mathbb{R}^{C \times \frac{C}{r}}$, $\mathbf{W}_2 \in \mathbb{R}^{\frac{C}{r} \times C}$ being weights of two fully-connected layers and the ReLU operator $\delta(\cdot)$. The parameter $r$ indicates the bottleneck in the channel excitation, which encodes the channel-wise dependencies. Foreshadowing some of our results, the best performance is obtained by $r=2$.} The dynamic range of the activations of $\hat{\mathbf{z}}$ are brought to the interval $[0, 1]$, passing it through a sigmoid layer $\sigma(\hat{\mathbf{z}})$. The resultant vector is used to recalibrate or excite $\mathbf{U}$ to
\begin{equation}
\hat{\mathbf{U}}_{cSE} = [\sigma(\hat{z_1})\mathbf{u}_1, \sigma(\hat{z_2})\mathbf{u}_2, \cdots , \sigma(\hat{z_{C}})\mathbf{u}_{C}].
\end{equation}
\noindent
The activation $\sigma(\hat{z}_i)$ indicates the importance of the $i^{th}$ channel, which is either scaled up or down. As the network learns, these activations are adaptively tuned to ignore less important channels and emphasize the important ones. The architecture of the block is illustrated in Fig.~\ref{fig:GA}(a).

\begin{figure*}[t]
\centering
\includegraphics[width=0.999\textwidth]{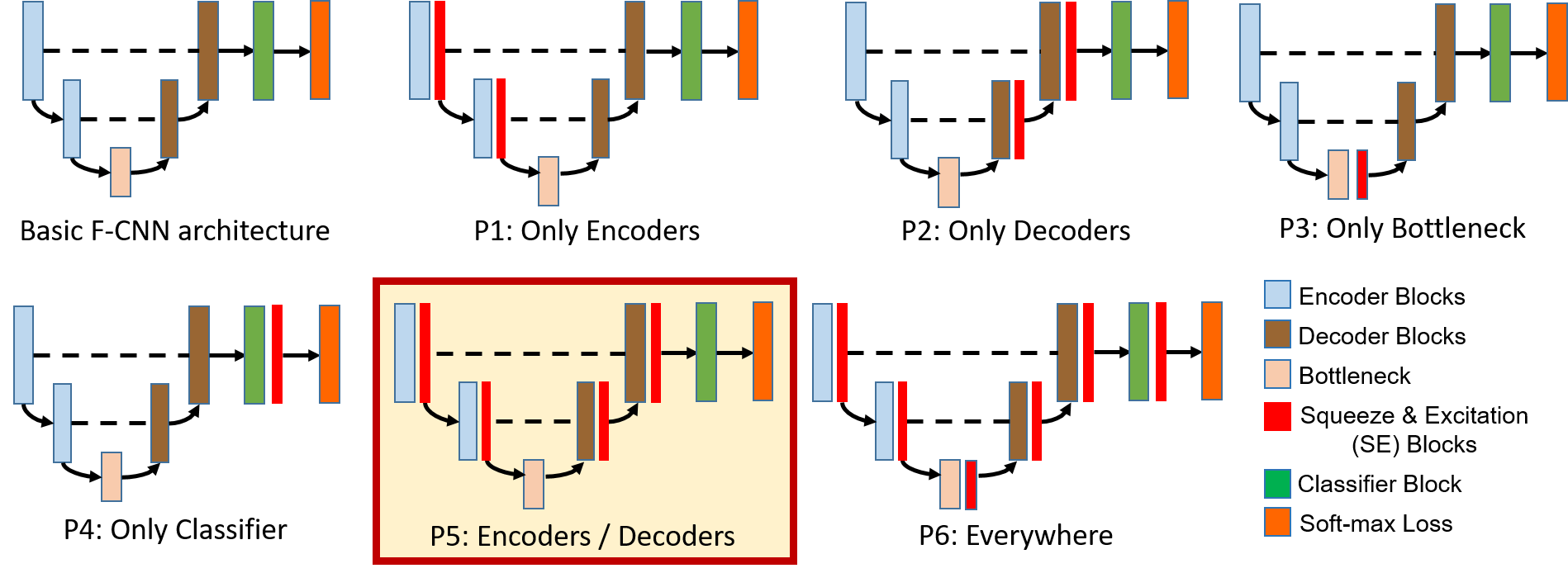}

\caption{Integration of SE blocks in encoder/decoder based F-CNN architectures. Illustration of the reference F-CNN architecture together with six possible integrations of SE blocks, described in Sec.~\ref{sec:position}. The recommended configuration P5 is highlighted, shown in Sec.~\ref{sec:posRes}}.
\label{fig:position}
\end{figure*}

\noindent
\subsection{Channel Squeeze and Spatial Excitation Block (sSE)}
We introduce the channel squeeze and spatial excitation block that squeezes the feature map $\mathbf{U}$ along the channels and excites spatially, which we consider important for fine-grained image segmentation. 
Here, we consider an alternative slicing of the input tensor $\mathbf{U} = [\mathbf{u}^{1,1}, \mathbf{u}^{1,2}, \cdots, \mathbf{u}^{i,j}, \cdots ,\mathbf{u}^{H,W}]$, where $\mathbf{u}^{i,j} \in \mathbb{R}^{1 \times 1 \times C}$ corresponding to the spatial location $(i,j)$ with $i \in \{1, 2, \cdots, H\}$ and $j \in \{1, 2, \cdots, W\}$.
The spatial squeeze operation is achieved through a convolution $\mathbf{q} = \mathbf{W}_{sq} \star \mathbf{U}$ with weight $\mathbf{W}_{sq} \in \mathbb{R}^{1 \times 1 \times C \times 1}$, generating a projection tensor $\mathbf{q} \in \mathbb{R}^{H \times W}$.
Each $q_{i,j}$ of the projection represents the linearly combined representation for all channels $C$ for a spatial location $(i,j)$. This projection is passed through a sigmoid layer $\sigma(.)$ to rescale activations to $[0,1]$, which is used to recalibrate or excite $\mathbf{U}$ spatially
\begin{equation}
\hat{\mathbf{U}}_{sSE} = [\sigma(q_{1,1})\mathbf{u}^{1,1}, \cdots, \sigma(q_{i,j})\mathbf{u}^{i,j}, \cdots, \sigma(q_{H,W})\mathbf{u}^{H,W}].
\end{equation}

\noindent
Each value $\sigma(q_{i,j})$ corresponds to the relative importance of a spatial information $(i,j)$ of a given feature map. 
This recalibration provides more importance to relevant spatial locations and ignores irrelevant ones. The architectural flow is shown in Fig.~\ref{fig:GA}(b). 

\noindent
\subsection{Spatial and Channel Squeeze \& Excitation Block (scSE)}
\label{sec:scSE}
Each of the above explained cSE and sSE blocks has its unique properties. 
The cSE blocks recalibrates the channels of by incorporating global spatial information. This global average pooling layers provides a receptive field of whole spatial extent at each stage of the F-CNN, aiding the segmentation pipeline. 
In contrast, the receptive field is not changed in sSE blocks as the channel squeeze is achieved by a $1\times1$ convolution layer.
Rather it behaves like an spatial attention map, indicating where the network should focus more to aid the segmentation.
We propose a combination of the complementary information from these two SE blocks, by concurrently recalibrating the input $\mathbf{U}$ spatially and channel-wise. 
The architecture of the combined scSE block is illustrated in Fig.~\ref{fig:GA}(c).
We explore four different strategies for the concurrent spatial and channel SE, $\hat{\mathbf{U}}_{scSE}$, in the following.

\noindent
(i) \textbf{Max-Out}: In this aggregation method, any location $(i,j,c)$ of the output feature map $\hat{\mathbf{U}}_{scSE}$ has the maximal activation  of $\hat{\mathbf{U}}_{cSE}$ and $\hat{\mathbf{U}}_{sSE}$. This corresponds to a location-wise max operator
\begin{equation}
\hat{\mathbf{U}}_{scSE}(i,j,c) = \max ( \hat{\mathbf{U}}_{cSE}(i,j,c),\hat{\mathbf{U}}_{sSE}(i,j,c) ).
\end{equation}
\noindent
The max-out layer enforces an element-wise competitiveness between the two SE blocks, similar to~\cite{liao2015competitive}. This provides a selective spatial  and channel excitation, such that the final segmentation is improved. 

\noindent
(ii) \textbf{Addition}: We add the two recalibrated feature maps $\hat{\mathbf{U}}_{cSE}$ and $\hat{\mathbf{U}}_{sSE}$ element-wise 
\begin{equation}
\hat{\mathbf{U}}_{scSE} = \hat{\mathbf{U}}_{cSE} + \hat{\mathbf{U}}_{sSE}.
\end{equation}
\noindent
This aggregation provides equal importance to the two sSE and cSE methods~\cite{roy2018concurrent}.

\noindent
(iii) \textbf{Multiplication}: We multiply the feature maps $\hat{\mathbf{U}}_{cSE}$ and $\hat{\mathbf{U}}_{sSE}$ element-wise 
\begin{equation}
\hat{\mathbf{U}}_{scSE}(i,j,c) = \hat{\mathbf{U}}_{cSE}(i,j,c) \times \hat{\mathbf{U}}_{sSE}(i,j,c).
\end{equation}
\noindent
Each location $(i,j,c)$ gets multiplied by both spatial and channel importance, each of which are scaled to $[0,1]$.

\noindent
(iv) \textbf{Concatenation}: We concatenate the two input responses, along the channel index, and pass it to the next encoder/decoder block
\begin{equation}
\hat{\mathbf{U}}_{scSE} = \mathtt{concat}(\hat{\mathbf{U}}_{cSE}, \hat{\mathbf{U}}_{sSE}).
\end{equation}
\noindent
Compared to the previously mentioned aggregation strategies, the advantage of this aggregation is that no information is lost. But on the downside, the number of channels of the output $\hat{\mathbf{U}}_{scSE}$ doubles, which in turn increases the model complexity as the subsequent convolutional layer must process feature maps with more channels.

\subsection{Position of SE Block in F-CNNs}
\label{sec:position}
One central question for integrating the proposed SE blocks in F-CNNs is their optimal position in the network to achieve the best performance. 
We explore six different  positions listed below:

\begin{enumerate}
\item[\textbf{P1:}] After encoder blocks.
\item[\textbf{P2:}] After decoder blocks.
\item[\textbf{P3:}] After bottleneck block.
\item[\textbf{P4:}] After classifier block.
\item[\textbf{P5:}] After encoder and decoder blocks.
\item[\textbf{P6:}] After encoders, decoders, bottleneck \& classifier.
\end{enumerate}

Fig.~\ref{fig:position} shows a sample encoder/decoder based F-CNN architecture, with all the defined positions of SE blocks  in P1 to P6.

\subsection{Model Complexity}
Let us consider an encoder/decoder block, with an output feature map of $C$ channels. 
Addition of a cSE block introduces $C^2$ new weights \revision{(assuming $r=2$)}, while an sSE block introduces $C$ weights. So, the increase in model complexity of an F-CNN with $h$ encoder/decoder blocks is
$\sum_{i=1}^h (C_i^2 + C_i)$,
where $C_i$ is the number of output channels for the $i^{th}$ encoder/decoder block. 
To give a concrete example, the U-Net in our experiments has about $2.1 \times 10^6$ parameters. The scSE block adds $3.3 \times 10^4$ parameters, which is an approximate increase by 1.5\%. 
Hence, SE blocks only increase overall network complexity by a very small fraction. 
\revision{In Tabel~\ref{tab:model_complexity}, we present the number of learnable parameters of the three F-CNN architectures (FC-DenseNet, SD-Net and U-Net), along with additional parameters due to inclusion of the cSE, sSE and scSE blocks. Most of the added model complexity is due to cSE. The sSE block causes only about $0.01\%$ increase in model complexity, which is negligible.}

\begin{table*}
\caption{\revision{Effect on model complexity for the addition of cSE, sSE and scSE blocks on the different F-CNN architectures. The table shows the number of parameters in each of the models, the additional parameters and the percentage increase in model complexity. }}
\centering
\revision{
\begin{tabular}{|c|c|c|c|c|}
    \hline
    & \textbf{Parameters} & \multicolumn{3}{c|}{\textbf{Additional Parameters (\% increase)}} \\
     Networks & No SE Block & + cSE Block & + sSE Block & + scSE Block  \\
    \hline
    \textbf{FC-DenseNets}\cite{densenet} & $3.1 \times 10^6$ & $3.2 \times 10^4(+1.03\%)$ & $512(+0.01\%)$ & $3.3 \times 10^4(+1.06\%)$  \\ 
    \textbf{SD-Net}\cite{ecb2017} & $1.8 \times 10^6$ & $3.2 \times 10^4 (+1.17\%)$ & $512(+0.02\%)$ & $3.3 \times 10^4(+1.18\%)$   \\ 
    \textbf{U-Net}\cite{Unet} & $2.1 \times 10^6$ & $3.2 \times 10^4(+1.52\%)$ & $512(+0.02\%)$ & $3.3 \times 10^4 (+ 1.57\%)$ \\ \hline
  \end{tabular}
  }
    \label{tab:model_complexity}
\end{table*}

\section{Experimental Setup}
\label{sec:exp}
We conducted extensive experiments to explore the impact of our proposed modules. 
We select three state-of-the-art F-CNN architectures, U-Net~\cite{Unet}, SD-Net~\cite{ecb2017} and FC-DenseNet~\cite{densenet}. All of the networks have an encoder/decoder based architecture. 

\subsection{Datatsets}
We use three datasets in our experiments.
(i) First, we address the task of segmenting MRI T1 brain scans into 27 cortical and sub-cortical structures. We use the Multi-Atlas Labelling Challenge (MALC) dataset~\cite{malc}, which is a part of OASIS~\cite{oasis}, with 15 scans for training and 15 scans for testing consistent to the challenge instructions. The main challenge associated with the dataset is the limited training data with severe class imbalance between the target structures. Manual segmentations for MALC were provided by Neuromorphometrics, Inc\footnote{http://Neuromorphometrics.com/}. Also, no pre-processing like skull-stripping, intensity re-normalization were performed. All the scans were re-sampled to isotropic resolution (1mm$^3$ voxel).
(ii) Second, we tackle the task of segmenting 10 organs on whole-body contrast enhanced CT (ceCT) scans. We use scans from the Visceral dataset~\cite{visceral}. We train on 65 scans from the silver corpus, and test on 20 scans with manual annotations from the gold corpus. The silver corpus was automatically labeled by fusing the results of multiple algorithms, yielding noisy labels. All the data were re-sampled to 2mm$^3$ voxel size. The main challenge associated with whole-body segmentation is the highly variable shape of the visceral organs and the capability to generalize when trained with noisy labels.
(iii) Third, we tackle the task of segmenting 7 retinal layers with accumulated fluid, on retinal OCT scans. We use the publicly available benchmark dataset from Duke University~\cite{chiu2015kernel}. The dataset consist of 110 B-scans from 10 Diabetic Macular Edema (DME) patients (i.e. 11 scans per subject). Scans from subjects 1-5 were used for training, and subjects 6-10 for testing, consistent with~\cite{roy2017relaynet,chiu2015kernel}. The main challenge with this task is the noisy nature of OCT scans, which makes segmentation of some thin retinal layers (5-6 pixels) and fluid very challenging. 
We use the Dice score for performance evaluation for all the datasets.

\subsection{Model Architecture and Learning}
In our experiments on the first two datasets, all of the three F-CNN architectures had $4$ encoder blocks, one bottleneck layer, $4$ decoder blocks and a classification layer at the end. 
The number of output feature maps in each of the blocks were kept constant to $64$. U-Net had $2$ padded $3\times3$ convolutional layers with ReLU in each of the block, without batch-norm layer~\cite{Unet}. SD-Net had only one $7\times7$ convolutional layer with batch-norm, terminating with ReLU layer in each of the block~\cite{ecb2017}. DenseNet had $2$ padded $5\times5$ convolutions with dense connections, terminating with another bottleneck $1\times1$ convolutional layer, similar to \cite{densenet}. On the third retinal OCT dataset, we use ReLayNet~\cite{roy2017relaynet} architecture, which is similar to SD-Net~\cite{ecb2017}. 

All the architectures operate in 2D, and training and testing was performed 2D slice-wise.
The logistic loss function was weighted with median frequency balancing~\cite{segnet} to compensate for the class imbalance. The learning rate was initially set to $0.01$ and decreased by one order after every 10 epochs. The momentum was set to $0.95$, weight decay constant to $10^{-4}$ and a mini batch size of $4$ was used. It must be noted that these settings are very commonly used. Optimization was performed using stochastic gradient descent. Training was continued till validation loss converged.
All the model and training settings were kept consistent with the original implementation.
All these experiments were conducted on an NVIDIA Titan Xp GPU with 12GB RAM.

\begin{table}[t]
\caption{Mean and standard deviation of the global Dice scores for SD-Net on MALC Dataset, with different aggregation strategies of sSE and cSE blocks in the concurrent scSE block.}
\centering
\begin{tabular}{|c|c|}
    \hline
    \textbf{Aggregation Strategy} & \textbf{Dice Score} \\ 
    \hline
    Max-Out & $\mathbf{0.867\pm0.082}$ \\
    Addition & $0.862\pm0.082$ \\
    Multiplication & $0.858\pm 0.035$ \\
    Concatenation & $\mathbf{0.868\pm 0.033}$ \\
    \hline
  \end{tabular}
  \label{tab:agg}
\end{table}
 
\begin{table*}
\caption{Mean and standard deviation of the global Dice scores for the different F-CNN models without and with cSE, sSE and scSE blocks on MALC and Visceral datasets.}
\centering
\begin{tabular}{|p{0.94in}|p{0.89in}|p{0.89in}|p{0.89in}|p{0.89in}|}
    \hline
    & \multicolumn{4}{c|}{\textbf{MALC Dataset}} \\
     Networks & No SE Block & + cSE Block & + sSE Block & + scSE Block  \\
    \hline
    \textbf{FC-DenseNets}\cite{densenet} & $0.842\pm0.058$ & $0.865\pm0.069$ & $0.876\pm0.061$ & $\mathbf{0.889}\pm0.065$  \\ 
    \textbf{SD-Net}\cite{ecb2017} & $0.771\pm0.150$ & $0.790\pm0.120$ & $0.860\pm0.062$ & $\mathbf{0.867}\pm0.082$   \\ 
    \textbf{U-Net}\cite{Unet} & $0.763\pm0.110$ & $0.825\pm0.063$ & $0.837\pm0.058$ & $\mathbf{0.851}\pm0.058$ \\ \hline
    & \multicolumn{4}{c|}{\textbf{Visceral Dataset}} \\
      Networks & No SE Block & + cSE Block & + sSE Block & + scSE Block  \\
    \hline
    \textbf{FC-DenseNets}\cite{densenet} & $0.892\pm0.068$ & $0.903\pm0.058$ & $0.912\pm0.056$ & $\mathbf{0.921}\pm0.078$  \\ 
    \textbf{SD-Net}\cite{ecb2017} & $0.871\pm0.064$ & $0.892\pm0.065$ & $0.901\pm0.057$ & $\mathbf{0.912}\pm0.026$   \\ 
    \textbf{U-Net}\cite{Unet} & $0.857\pm0.106$ & $0.865\pm0.086$ & $0.872\pm0.080$ & $\mathbf{0.887}\pm0.028$ \\ \hline
  \end{tabular}
    \label{tab:res}
\end{table*}

\section{Results and Discussion}
\label{sec:dis}

In this section, we present  quantitative and qualitative results of our experiments on the aforementioned three datasets. Further, we investigate  the best aggregation scheme for scSE, optimal position of the blocks within F-CNNs, and the dynamics of spatial excitation (sSE) during the training process.

\subsection{scSE Aggregation Strategies}
Here we investigate into the best aggregation strategy of sSE and cSE, among the four possibilities as described in Sec.~\ref{sec:scSE}. We present results for  SD-Net on MALC dataset for brain segmentation. The average Dice scores for all methods are reported in Tab.~\ref{tab:agg}. We observe that all  aggregation schemes boost the segmentation performance. Using max-out and concat  provides the best performance. As concat  aggregation increases the model complexity (increases the number of channels of the output feature map for each block), max-out provides the best trade-off between performance and model complexity.  
\revision{An intuitive explanation behind the superior performance of the max-out based aggregation is its ability to induce element-wise selectivity by making both of the excitations compete. This concept was previously used in classification where kernels with multiple scales where made to compete using max-out, instead of concatenation, providing similar performance with reduced model complexity~\cite{liao2015competitive}.}
In all the following experiments, we use max-out based aggregation for scSE blocks.

\subsection{Position of SE Blocks}
\label{sec:posRes}
In this section, we investigate different positions to use the scSE blocks within F-CNN architectures. We introduced six possible configurations in Sec.~\ref{sec:position}. We select SD-Net as architecture and MALC dataset for our experiments. We added the scSE blocks as per P1-P6 within SD-Net and reported the mean and standard deviation for the global Dice score on MALC test data in Tab.~\ref{tab:position}. Firstly, we observe that scSE blocks lead to a clear improvement of segmentation quality at every position (P1-P6) of the network. The effect is more prominent in encoders (P1) and decoders (P2), in comparison to bottleneck (P3) and classifier (P4). Combined configurations P5 and P6 exhibit similar performance. Out of these two, we choose P5 over P6, as it adds less complexity to the overall model.  In all the next experiments, we add  SE blocks as per P5, after every encoders and decoders.

\begin{table}[h]
\caption{Mean and standard deviation of the global Dice scores for SD-Net on MALC dataset with scSE blocks at different positions.}
\centering
\begin{tabular}{|c|c|}
    \hline
    \textbf{Position} & \textbf{Dice Score} \\ 
    \hline
    No scSE Block & $0.771\pm0.150$ \\
    P1 & $0.858\pm0.033$ \\
    P2 & $0.832\pm 0.032$ \\
    P3 & $0.821\pm0.032$ \\
    P4 & $0.803\pm0.034$ \\
    P5 & $\mathbf{0.867}\pm0.082$ \\
    P6 & $\mathbf{0.867}\pm0.036$ \\
    \hline
  \end{tabular}
  \label{tab:position}
\end{table}

\revision{
\subsection{Sensitivity to parameter $r$ in scSE}
One hyper-parameter of our proposed modules is the parameter $r$ the cSE block, which indicates the bottleneck in the channel excitation. In \cite{SE2017}, this was set to a value of $16$ for the task of image classification. To find its optimal setting in our application, we perform experiments with different values of $r = \{ 2, 4, 8, 16 \}$. We select SD-Net as our architecture and MALC dataset for evaluation purposes. The mean and standard deviation in global Dice score on the test data with different values of $r$ are reported in Tab.~\ref{tab:sensitivity_r}. We observe that as $r$ increases from $2$ to $16$, the global Dice score decreases; with $r = \{ 8, 16 \}$ it exhibits the same performance as using sSE only. As indicated earlier, the number of channels in the feature maps after the convolutions are $64$. Using $r = \{ 8, 16 \}$ reduces the bottleneck to $8$ or $4$ nodes respectively, which reduces the effect of channel excitation. We use $r = 2$ for all our experiments. }

\begin{table}[h]

\caption{\revision{Mean and standard deviation of the global Dice scores for SD-Net on MALC dataset with scSE blocks with different values of hyper-parameter $r$.}}
\centering
\revision{
\begin{tabular}{|c|c|}
    \hline
     & \textbf{Dice Score} \\ 
    \hline
    $r = 2$ & $\mathbf{0.867\pm0.082}$ \\
    $r = 4$ & $0.861\pm0.031$ \\
    $r = 8$ & $0.860\pm 0.030$ \\
    $r = 16$ & $0.860\pm0.026$ \\
    \hline
  \end{tabular}
  }
  \label{tab:sensitivity_r}
\end{table}

\revision{
\subsection{Position of skip connection}
Skip connections are a fundamental design choice within modern F-CNNs. In this section, we investigate the relative positions between skip connections and scSE blocks. One possibility is indicated in Fig.~\ref{fig:position}: the output of the encoder blocks are first re-calibrated using scSE blocks, and the re-calibrated feature map is used for the skip connection to the corresponding decoder with similar resolution (configuration-1). Another possible configuration would be to use the un-calibrated feature map before scSE block for skip connection from encoder to decoder (configuration 2). We performed experiments with both of these experiments on SD-Net on MALC dataset. Configuration 1 achieved a global Dice score of $0.867\pm0.082$ on test set, while configuration 2 achieved the score $0.869\pm0.026$. The performance is almost similar and the difference in not statistically significant ($p>0.05$, Wilcoxon signed-rank). This indicates that both configurations are equally effective and any one could be used.
}

\subsection{Whole Brain Segmentation}
Tab.~\ref{tab:res} presents the results of whole brain segmentation on the 15 MALC testing images with the three F-CNNs models (U-Net, SD-Net and FC-DenseNet) using cSE, sSE and scSE blocks. Comparing along the columns, we observe that the mean Dice score increases with inclusion of any of the SE blocks by a statistically significant margin ($p \le 0.001$, Wilcoxon signed-rank), in comparison to the reference architecture. This is consistent for all the F-CNNs, which indicates that  SE blocks aid in better representation learning for segmentation. 
Also, we observe that spatial excitation (sSE) is more effective in comparison to channel excitation (cSE) for segmentation, while its combination (scSE) provides the best performance. The increase in global Dice score using scSE blocks in comparison to the normal version is around 4\% for FC-DenseNets, and around 8-9\% for U-Net and SD-Net. Such increase in performance is striking, given the difficulty associated with such a task. We further observed that the increase is more profound in U-Net and SD-Net, as its reference architectures failed to segment some very small structures (3rd ventricle, amygdala). This was corrected by SE blocks. 

\begin{figure*}[t]
\centering
\includegraphics[width=0.98\textwidth]{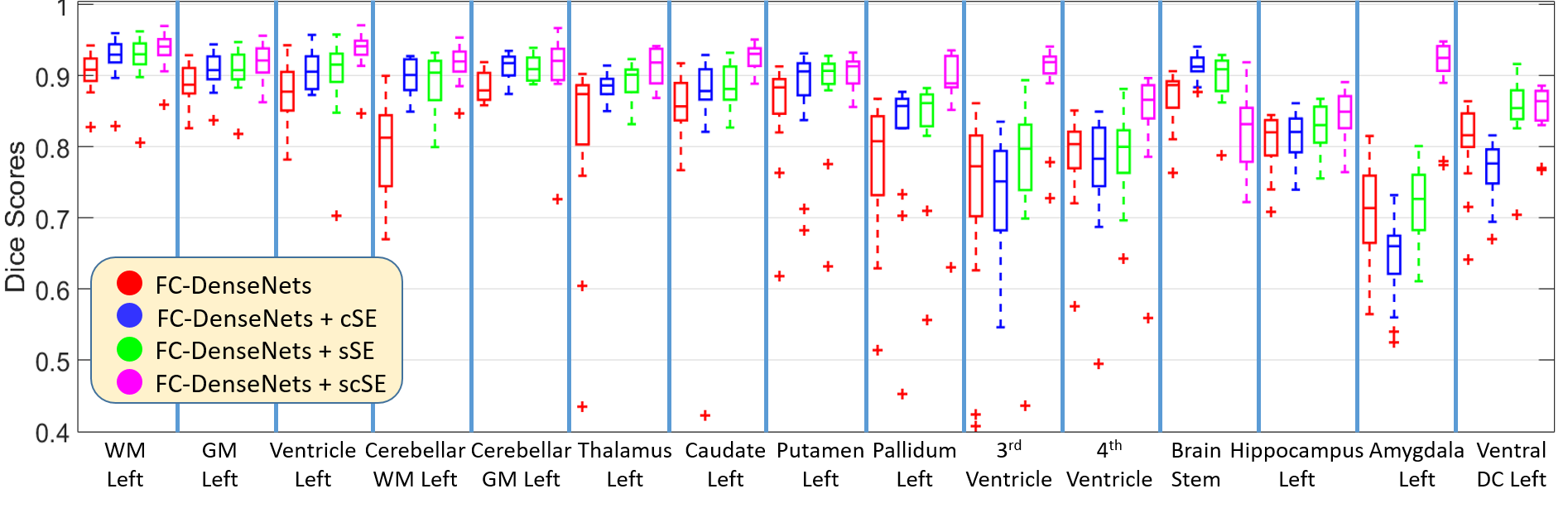}

\caption{Boxplot of Dice scores for all brain structures on the left hemisphere (due to space constraints), using DenseNets on MALC dataset, without and with proposed cSE, sSE, scSE blocks. Grey and white matter are abbreviated as GM and WM, respectively. Center-lines indicate the median, boxes extend to the 25th and 75th percentiles, and the whiskers reach to the most extreme values not considered outliers (indicated by crosses).}
\label{fig:plotBrain}
\end{figure*}

Fig.~\ref{fig:plotBrain} shows structure-wise Dice scores of FC-DenseNet, with cSE, sSE and scSE blocks. 
We present only the structures of left hemisphere due to space constraints. Here, we observe that sSE and scSE outperform the reference model  for almost all structures. 
The cSE model outperforms the reference model in most structures except some challenging structures like 3rd/4th ventricles, amygdala and ventral DC, where its performance degrades. One possible explanation could be the small size of these structures for channel excitation. This was compensated by the attention mechanism of the spatial excitation. 

Fig.~\ref{fig:result} (a-d) presents qualitative segmentation results with the MRI T1 input scan, ground truth annotations, FC-DenseNet segmentation along with the proposed FC-DenseNet+scSE segmentation, respectively. 
We highlight a ROI with a white box and red arrow, indicating the  left putamen, which is under segmented using  FC-DenseNet (Fig.~\ref{fig:result}(c)), but the segmentation improves with the inclusion of the scSE block (Fig.~\ref{fig:result}(d)).


\begin{figure*}[t]
\centering
\includegraphics[width=0.80\textwidth]{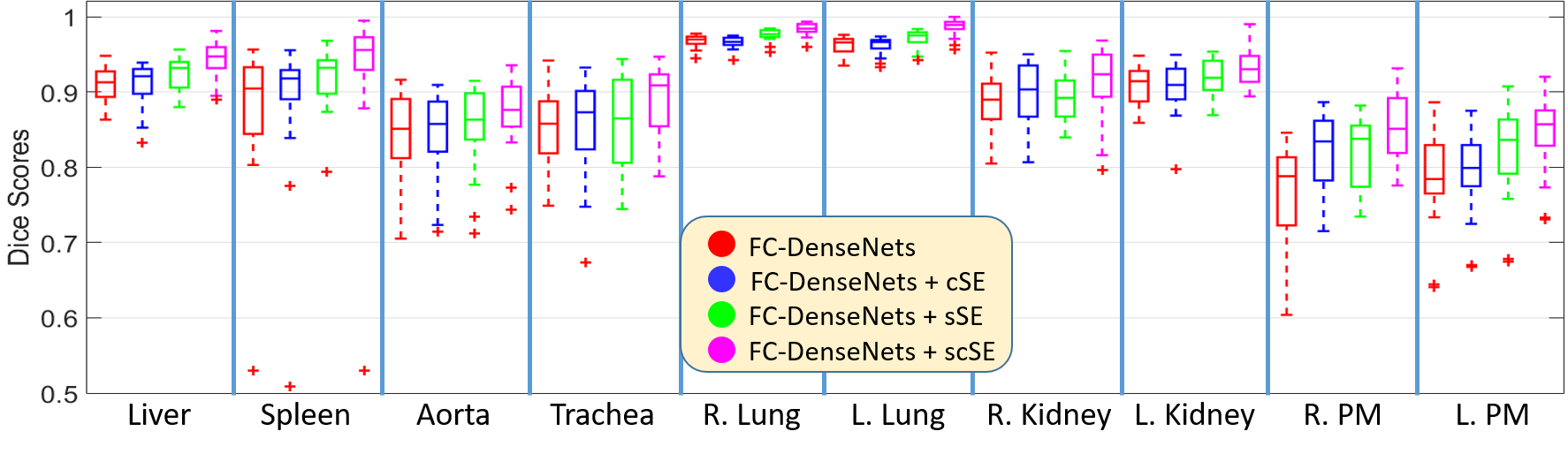}

\caption{Structure-wise Dice performance of DenseNets on Visceral dataset, without and with proposed cSE, sSE, scSE blocks. Left and right are indicated as L. and R. Psoas major muscle is abbreviated as PM. Center-lines indicate the median, boxes extend to the 25th and 75th percentiles, and the whiskers reach to the most extreme values not considered outliers (indicated by crosses).}
\label{fig:plotBody}
\end{figure*}

\subsection{Whole Body Segmentation}
\begin{figure*}[t]
\centering
\includegraphics[width=0.8\textwidth]{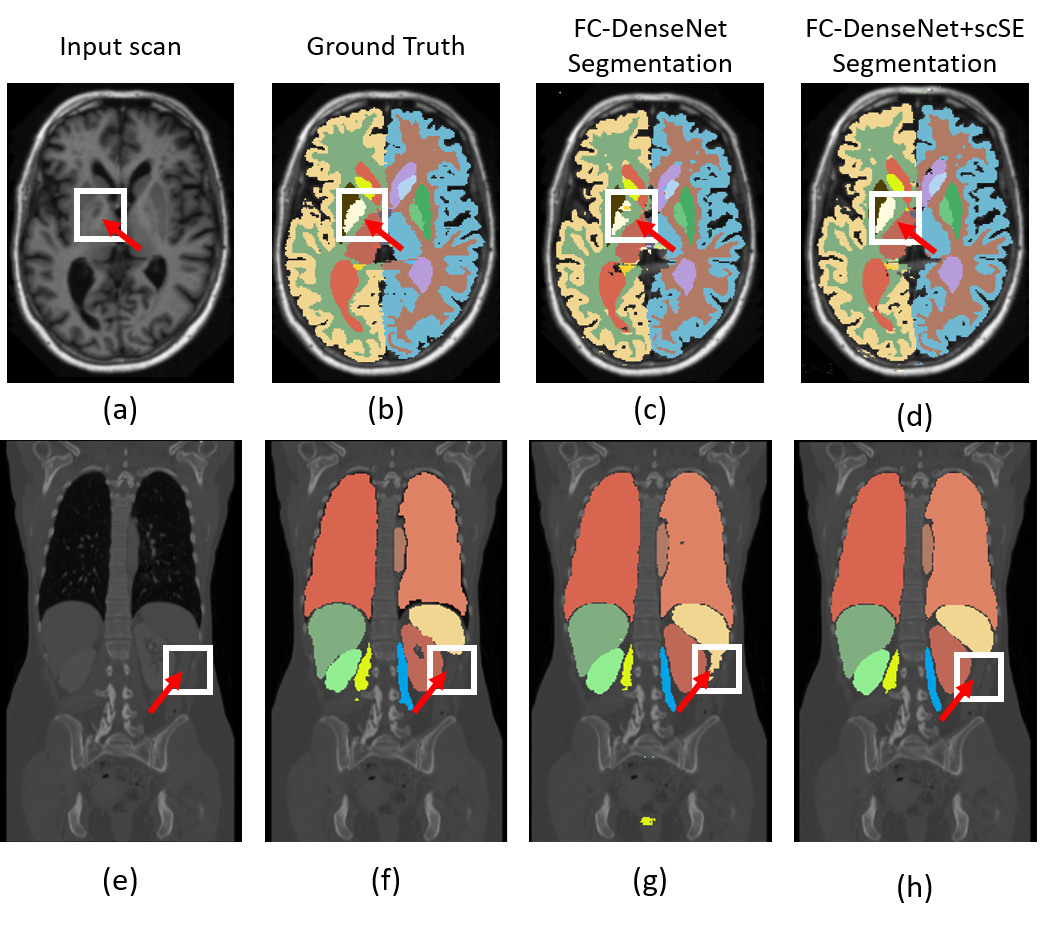}

\caption{Input scan, ground truth annotations, DenseNet segmentation and DenseNet+scSE segmentation for both whole-brain MRI T1 (a-d) and whole-body ceCT (e-h) are shown.  ROIs are indicated by white box and red arrow highlighting regions where the scSE block improved the segmentation, for both applications.}
\label{fig:result}
\end{figure*}

Tab.~\ref{tab:res} also presents  the results for abdominal organ segmentation in 20 test ceCT scans of the Visceral dataset. Comparing along the columns, we observe a similar trend of increase in performance by addition of the SE blocks ($p\le0.05$, Wilcoxon signed rank) as for whole brain segmentation. Also, the effect of sSE is more prominent than cSE, while scSE is the best. Comparing across rows, addition of scSE blocks provided an increase of 3\%, 4\% and 3\% in FC-DenseNet, SD-Net and U-Net, respectively. 
Compared to brain segmentation, the increase in Dice score is less in this dataset. 
This is because the segmentation task is easier than for brain segmentation, leading to higher Dice scores in the baseline architectures. 
Fig.~\ref{fig:plotBody} presents  organ-wise Dice scores for FC-DenseNets with the addition of cSE, sSE and scSE blocks. We observe that the addition of scSE provides a consistent increase in Dice scores for all organs. We present a test ceCT scan, with manual annotations, FC-DenseNet segmentations without and with scSE block in Fig.~\ref{fig:result} (e-h). Here we highlight a region with a white box and a red colored arrow, where the spleen gets over segmented using DenseNets, while it gets rectified by addition of scSE blocks.

\subsection{Retinal Layer and Fluid Segmentation}
In this section, we present the results for segmentation of retinal OCT B-scans into 7 retinal layers and accumulated fluid. 
We recently introduced ReLayNet~\cite{roy2017relaynet}, which achieved the best performance on the benchmark dataset. We add scSE blocks after every encoder/decoder block in ReLayNet. The training procedure was kept consistent with the original implementation in~\cite{roy2017relaynet}\footnote{https://github.com/abhi4ssj/ReLayNet}. We present the class-wise and overall Dice scores in Tab.~\ref{tab:relaynet}. We observe that there is a consistent increase in Dice score by 1-2\% in most of the classes by adding scSE blocks.

\begin{table}[h]
\caption{Class-wise mean Dice scores for retinal SD-OCT segmentation, using ReLayNet~\cite{roy2017relaynet}, with and without scSE Blocks.}
\centering
\begin{tabular}{|c|c|c|}
    \hline
    \textbf{Classes} & \textbf{ReLayNet} & \textbf{ReLayNet + scSE Block} \\ 
    \hline
    ILM & $0.90$ & $\mathbf{0.91}$ \\
    NFL-IFL & $0.94$ & $\mathbf{0.95}$ \\
    INL & $0.87$ & $\mathbf{0.88}$ \\
    OPL & $0.84$ & $\mathbf{0.86}$ \\
    ONL-ISM & $\mathbf{0.93}$ & $\mathbf{0.93}$ \\
    ISE & $\mathbf{0.92}$ & $\mathbf{0.92}$ \\
    OS-RPE & $0.90$ & $\mathbf{0.91}$ \\
    Fluid & $0.77$ & $\mathbf{0.79}$ \\ \hline
    Overall & $0.883$ & $\mathbf{0.893}$ \\
    \hline
  \end{tabular}
  \label{tab:relaynet}
\end{table}

\begin{figure*}[h]
\centering
\includegraphics[width=0.98\textwidth]{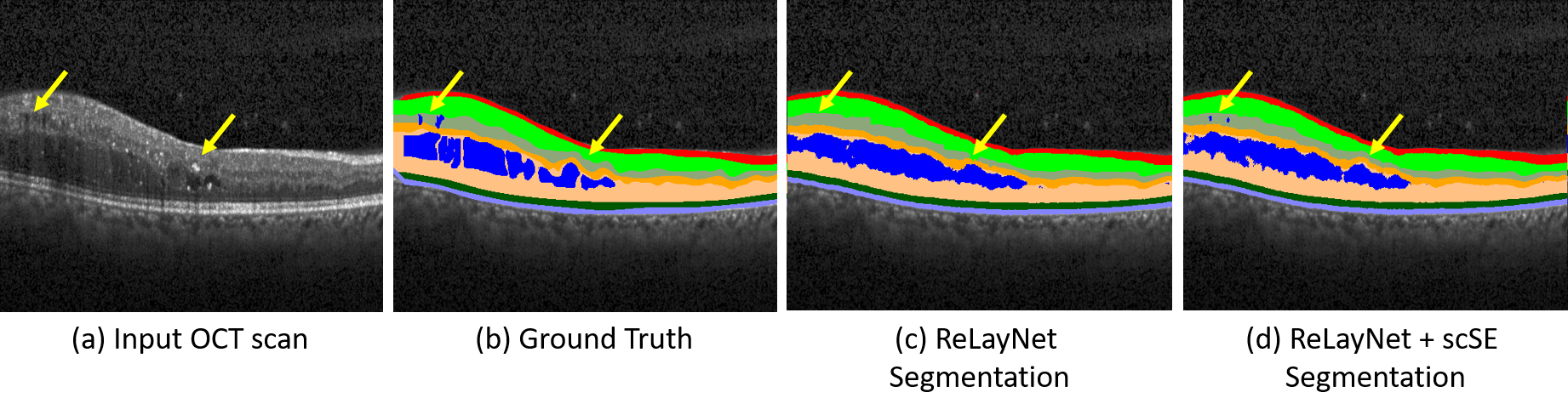}
\caption{Input retinal OCT B-scan (a), manual annotations of layer and fluid class (b), segmentation with ReLayNet (c), and segmentation using ReLayNet with scSE blocks (d). We highlight two regions with yellow arrows, where segmentation is improved by using the scSE blocks.}
\label{fig:relaynet}
\end{figure*}

Further, we present a test OCT scan (with DME), with manual annotation, ReLayNet predictions without and with scSE blocks in Fig.~\ref{fig:relaynet} (a-d) respectively. We highlight two region of interests with yellow arrows. The arrow to the left indicates two very small pools of fluid masses with INL region, which is visually imperceptible. The addition of scSE blocks managed to detect the fluid masses, whereas the normal ReLayNet failed to do so. Another arrow at the middle indicates a region in INL, where there is a bright imaging artifact. This resulted in a discontinuous prediction using ReLayNet, which is rectified by adding  scSE blocks.

\subsection{scSE vs. Additional Encoder/Decoder Block}
One could argue that the increase in performance by adding scSE blocks may be because of: (i) added model complexity and (ii) increased receptive field (average global pooling). 
A naive way to incorporate these properties within an F-CNN architecture is by adding an extra encoder/decoder block.
Tab.~\ref{tab:scSEvLayer} shows the results of SD-Net on MALC for the reference architecture, which consists of 3 encoder/decoder blocks, the addition of another encoder/decoder block, and the addition of scSE blocks. 
We observe that adding an extra encoder/decoder improved the Dice score by 5\% but with an increase of 8\% in model complexity. Whereas, adding scSE increases the Dice score by 9\%, while only increasing model complexity by 1.5\%. This substantiates that the scSE blocks has unique properties, which cannot be achieved by adding more convolutional layers.

\begin{table}[h]
\caption{Mean and standard deviation of the global Dice scores on MALC for SD-Net,  for SD-Net with additional encoder/decoder block, and for SD-Net with scSE, together with the added model complexity.}
\centering
\begin{tabular}{|c|c|c|}
    \hline
    \textbf{Model} & \textbf{Dice Score} & \textbf{Complexity} \\ 
    \hline
    SD-Net & $0.771\pm0.150$ & - \\
    SD-Net + enc/dec block & $0.823\pm0.031$ & +33.2\% \\
    SD-Net + scSE & $\mathbf{0.867\pm0.082}$ & +1.8\%\\
    \hline
  \end{tabular}
  \label{tab:scSEvLayer}
\end{table}

\subsection{Dynamics of Spatial Excitation}

\begin{figure*}[t]
\centering
\includegraphics[width=0.9\textwidth]{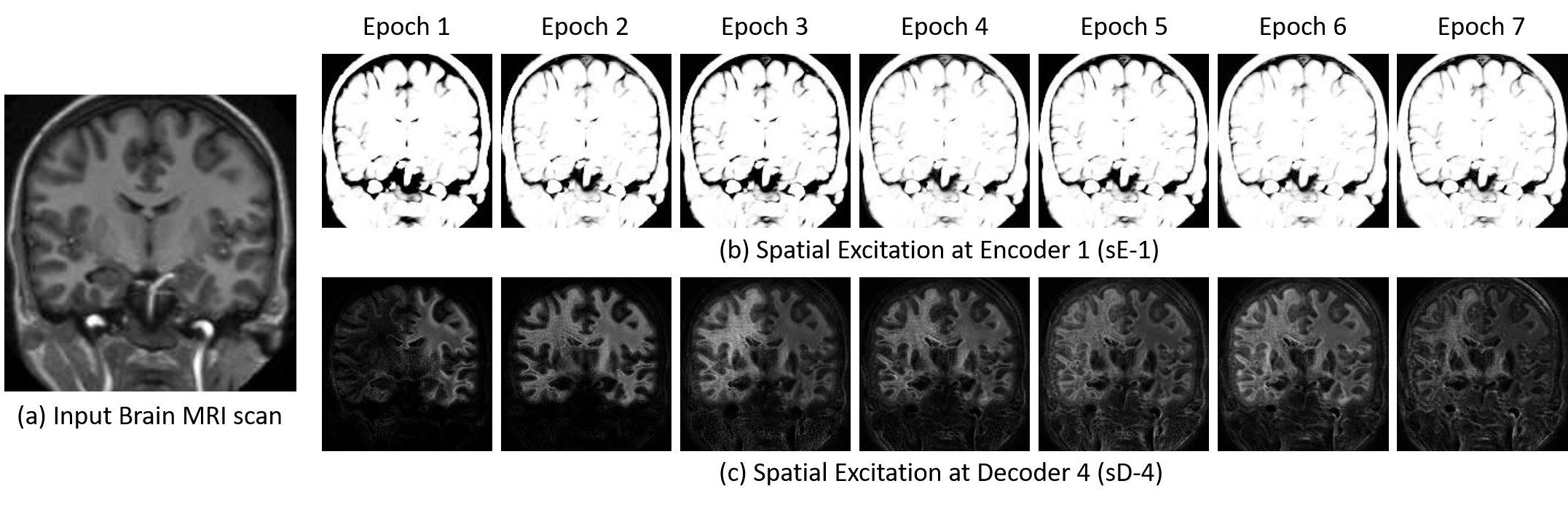}
\caption{Dynamics of  spatial excitation during  training. (a) shows the input brain MRI scan, while (b) and (c) shows the spatial activations for the first encoder (sE-1) and the spatial activation maps for last decoder (sD-4),  respectively, for epochs 1 to 7.}
\label{fig:dynamics}
\end{figure*}

After empirically demonstrating the effectiveness of SE blocks in F-CNN segmentation, we now try to understand the dynamics of squeeze \& excite during training. 
We use FC-DenseNets, save the network after each epoch in training and only include spatial SE blocks to reduce complexity; channel excitation was previously analyzed in~\cite{SE2017}. 
It is generally very difficult to interpret internal representations of a deep neural network, where we focus on the spatial activation maps of the first encoder (sE-1) and last decoder (sD-4), as they have the same spatial dimensions as the input scan. 
Note that spatial activation maps are element-wise multiplied with feature maps. 
Fig.~\ref{fig:dynamics} shows spatial activation maps at sE-1 and sD-4 for epochs 1 to 7, for an MRI brain scan. 

%

Activation maps in sE-1 mainly show foreground vs. background based distinction. 
As this is one of the shallow layers, with lower orders of features, the activations are not class specific. In brain MRI scans, we observe that skull is also highlighted, although it is part of the background class. This indicates that the network uses the skull as reference to establish relative spatial locations to brain tissues. 
Also, sE-1 maps don't change much over the epochs as this is one of the first layers in the network and already learned during the first epoch. 

For sD-4 activation maps, we observe a more dynamic behavior with clear changes of activation during epochs. 
For brain MRI, we see a highlight on the left white matter after the first epoch, while the right white matter is more highlighted after the third epoch. For other epochs, the activation across hemispheres is more balanced. 
It is interesting to observe that the network has already roughly learned the white matter in the first epoch, as indicated by the activation following it that closely. 
From the fifth epoch, cerebellar structures get highlighted. 
Overall, we note a certain similarity of activation maps in later epochs to the input scan with a focus on boundaries. 


\section{Conclusion}
\label{sec:conc}
We proposed to integrate the `squeeze \& excitation'  blocks within F-CNNs, to recalibrate intermediate feature maps, for image segmentation. 
We introduced the \emph{spatial} `squeeze \& excitation', which outperforms the previously proposed channel-wise `squeeze \& excitation', for segmentation tasks. Towards this end, we propose a combine these spatial and channel `squeeze \& excitation' within a single block.
In our extensive set of validations on three different F-CNN architectures and three different segmentation applications, we demonstrate that SE blocks yield a consistent improvement in segmentation performance. 
Hence, recalibration with SE blocks seems to be a fairly generic concept to boost performance in F-CNNs. 
Strikingly, the substantial increase in segmentation accuracy comes with a negligible increase in model complexity. 
With the seamless integration, we believe that squeeze \& excitation can be a crucial component for F-CNNs in many medical applications.

\bibliographystyle{IEEEtran}

\bibliography{jab_bib}

\end{document}